\def\year{2019}\relax
\documentclass[letterpaper]{article} 
\usepackage{aaai19}  
\usepackage{times}  
\usepackage{helvet}  
\usepackage{courier}  
\usepackage{url}  
\usepackage{graphicx}  
\usepackage{amsmath,amssymb,amsfonts}
\usepackage{algorithmic}
\usepackage{caption,graphicx}
\usepackage{esvect}
\usepackage{float}

\usepackage{dblfloatfix}

\usepackage{subcaption}

\begin{document}
%
\title{Coupled IGMM-GANs for deep multimodal anomaly detection\\ in human mobility data\\
\author{Kathryn Gray\\ Dept of Applied Mathematics \\ Dept of Computer Science \\ University of Colorado at Boulder \\Boulder, Colorado \\ kathryn.gray@colorado.edu
\And Daniel Smolyak\\Dept of Computer Science\\University of Maryland, College Park\\ College Park, Maryland\\dsmolyak@umd.edu \And Sarkhan Badirli \and George Mohler\\
Dept of Computer \& Information Science\\ IUPUI\\ Indianapolis, Indiana \\ \{sbadirli, gmohler\}@iu.edu}}

\maketitle

\begin{abstract}
Detecting anomalous activity in human mobility data has a number of applications including road hazard sensing, telematic based insurance, and fraud detection in taxi services and ride sharing.  In this paper we address two challenges that arise in the study of anomalous human trajectories: 1) a lack of ground truth data on what defines an anomaly and 2) the dependence of existing methods on significant pre-processing and feature engineering.  While generative adversarial networks seem like a natural fit for addressing these challenges, we find that existing GAN based anomaly detection algorithms perform poorly due to their inability to handle multimodal patterns.  For this purpose we introduce an infinite Gaussian mixture model coupled with (bi-directional) generative adversarial networks, IGMM-GAN, that is able to generate synthetic, yet realistic, human mobility data and simultaneously facilitates multimodal anomaly detection.  Through estimation of a generative probability density on the space of human trajectories, we are able to generate realistic synthetic datasets that can be used to benchmark existing anomaly detection methods.  The estimated multimodal density also allows for a natural definition of outlier that we use for detecting anomalous trajectories.  We illustrate our methodology and its improvement over existing GAN anomaly detection on several human mobility datasets, along with MNIST.  

\end{abstract}

\section{Introduction}

Human mobility data has become widely available through mobile, wearable, and in-vehicle sensing and has motivated a variety of machine learning tasks.  One area of research focuses on activity detection and classification \cite{shoaib2015survey}, where GPS or other sensor time series are used to classify trajectories (e.g. walking vs. biking).  While in activity detection an exhaustive set of classes is known, the related task of anomaly detection in mobility data \cite{zhang2011ibat} involves discovering samples that may belong to a new class or are outliers in the dataset.  Finding anomalies in GPS trajectory data has a wide range of applications including determining anomalous traffic patterns \cite{pan2013crowd}, finding taxi drivers who commit fraud \cite{6450098}, and driver finger printing and hidden driver detection in usage based insurance \cite{enev2016automobile,ramaiah2016deep}.

In the majority of activity and anomaly detection studies, significant pre-processing and feature engineering is used prior to classification or trajectory similarity comparisons.  While end-to-end deep learning has been recently applied to activity recognition \cite{ramaiah2016deep}, to our knowledge no work to date has applied deep learning to anomaly detection in human mobility data.  Furthermore, unlike activity recognition where ground-truth data is available, anomaly definition is often vague and subjective.  With no ground truth datasets available, it is difficult to compare benchmark models available for detecting anomalies in trajectory and other sensor data.

\begin{figure*}
\centering
\includegraphics[scale=0.65]{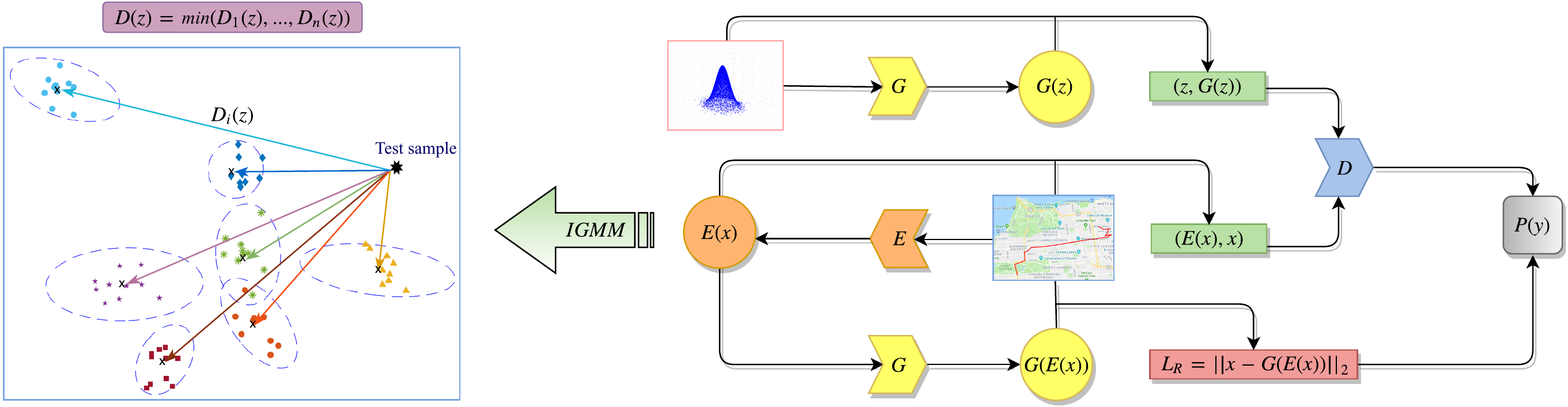}
\caption{The IGMM-GAN architecture.  The generator network learns to transform Gaussian samples into synthetic trips, while the discriminator network learns to distinguish real from fake trips.  Simultaneously, and encoder network learns the inverse mapping of the generator for trip embedding in the latent space.  Finally, a multimodal Mahalanobis distance metric from the IGMM is used to detect outliers in unseen test data.   }
\label{figure_diagram}
\end{figure*}

To overcome these challenges, we propose using Generative Adversarial Networks (GAN) \cite{goodfellow2014generative} that can simultaneously generate realistic trajectory data as well as detect anomalies.  In Figure \ref{figure_diagram}, we provide an overview of our coupled IGMM-GAN model.  We use a Bidirectional GAN (BiGAN) that learns an encoder in addition to a generator neural network for transforming trajectory data into a latent space where outliers may be detected.  Unlike previous unimodal GAN based anomaly detection \cite{schlegl2017unsupervised,zenati2018efficient}, we use an Infinite Gaussian Mixture Model to detect anomalies in the latent space through a multi-modal Mahalanobis metric.  We find this approach to significantly improve the accuracy of previous GAN based anomaly detection algorithms.  

The outline of the paper is as follows.  In Section 2, we review related work on anomaly detection in trajectory data and generative adversarial networks.  In Section 3, we provide details on the IGMM-GAN model.  In Section 4, we present experimental results applying our model to MNIST and several human trajectory datasets.  We compare AUC scores of the IGMM-GAN against several recently proposed GAN based anomaly detection algorithms and also provide a qualitative analysis of the generated synthetic trajectories and anomalies of the IGMM-GAN.  

\section{Related Work}

In this section we review previous literature on methods of detecting anomalies in GPS data and the use of GANs for synthetic data generation, embedding, and anomaly detection.

\subsection{GPS Anomalies}

Previous studies on anomaly detection in GPS trajectory data have focused both on the detection of city-wide traffic events and determining abnormal driving patterns on the level of an individual driver.
On the more global level, \cite{donovan2015using} used GPS data from New York City cab drivers to determine what effect natural disasters and other large scale disruptions had on the driving times in the city. In their work, traffic levels between regions are determined through the analysis of origin/destination pairs for taxi trips, allowing extrapolations to be made as to the traffic levels of the city as a whole.
Combining GPS data and Twitter data, \cite{pan2013crowd} were able to identify both the presence of increased traffic and its possible causes, such as accidents or weather. To find traffic anomalies, they analyzed road networks and driver trajectories as part of a graph, looking for time periods and routes where activity deviated from historical patterns. \cite{pang2013detection} also check for deviations in historical patterns, but instead use statistical spatio-temporal models of the number of taxis in a given region. Anomalies are found simply by looking for data points that are considered outliers within the statistical model of each region.

At the individual level, \cite{zhang2011ibat} developed the Isolation Based Anomalous Trajectory (iBAT) detection method, which compares a driver's single trajectory to other trajectories with the same origin and destination. Dividing the space into a grid, iBAT calculates an anomaly score based on the similarity of the new route, in terms of the number of matching grid spaces.  The related Isolation Based Anomalous Online Trajectory (iBOAT) detection method \cite{chen2013iboat}, that consists of examining sub-trajectories through the lens of an adaptive working window, is able to complete similar calculations in real time and with more specificity, identifying precisely which part of a trajectory is anomalous. These two methods are able to detect not only increases in traffic, but also instances of fraud, where taxi drivers would deliberately take their passengers on longer routes in order to increase their earnings. \cite{bu2009efficient} monitor for anomalies over continuous data streams as well, but instead use Euclidean distance to compare the trajectory under analysis to past and future trajectories. Much of their work focuses on using local clustering and other methods of optimization to complete these comparisons efficiently.  Our goal in this paper is to develop an alternative, end-to-end learning framework, that does not require pre-processing trips on to grids or costly trip similarity scores.  Instead, trips will be embedded in a Euclidiean space where a multi-modal Mahalanobis distance is used to flag outliers.

\subsection{GANs for synthetic data generation, embedding and anomaly detection}

There have been several recent applications of GANs to synthetic data generation.  \cite{76a8445ca17f4bee9222efbac37e1dd6} applied GANs to generate and analyze data on the activity patterns of neurons, with the goal of creating data exhibiting the same statistics as the real data.  To mitigate imbalanced data, \cite{salehinejad2017generalization} used GANs to generate synthetic chest x-ray images.  The synthetic data augmented samples of rare conditions and training with the generated data was shown to improve classifiers designed to identify these rare conditions. 

\cite{gupta2018social} generated movement trajectories based on socially acceptable behavior.  These behaviors included passing and meeting people while walking and included elements such as speed and direction.  GANs were then used to predict possible trajectories given a past trajectory. Latent representations of the trajectories were also used to determine direction and speed.

The closest past work on human mobility and deep data generation to ours is \cite{DBLP:journals/corr/AlzantotCS17}, who used a GAN-like architecture to generate synethetic acceleration data.  Long-Short-Term-Memory networks coupled with mixture density networks were used to generate the acceleration time-series.  However, the authors did not use a complete GAN architecture, in particular they did not use a discriminator to help train the generator and did not use an encoder for latent representations.  The paper also focused solely on acceleration data, not GPS trajectories.

Several recent studies \cite{schlegl2017unsupervised,zenati2018efficient} have successfully applied GANs for the purpose of anomaly detection making use of Bidirectional GAN (BiGAN) \cite{donahue2016adversarial}.  These methods have fared favorably in anomaly detection compared to other deep embedding methods such as variational auto-encoders.  However, we have found the existing GAN based anomaly detection methods (GANomaly and Efficient GAN Anomaly Detection) to have difficulties when the data is multimodal.  In the following sections we show that they can be significantyly improved with a multimodal framework.

\section{Methodology}

\subsection{GANs}

GANs, first proposed in \cite{goodfellow2014generative}, consist of a generator (G) network and a discriminator (D) network: the two follow the below minimax game, where the generator tries to minimize the $\log(1-D(G(z)))$ term and the discriminator tries to maximize the $\log(D(x))$ term.
\begin{align*}
    \max\limits_{D}\min\limits_{G} V(D, G) = \mathbb{E}_{x\sim p_{data}(x)}[\log D(x)]\\ + \mathbb{E}_{z\sim p_z(z)}[\log (1-D(G(z)))]
\end{align*}

The discriminator network improves the loss when it classifies a sample $x$ correctly and $D(x)$ is the probability that $x$ is real rather than generated data.  Meanwhile, the generator network maps Gaussian samples $z$ into synthetic data samples $G(z)$ (e.g. image or GPS trajectory).  The generator attempts to minimize the discriminator loss by generating a fake sample $G(z)$ such that the discriminator labels the sample as real (hence the $1-D(G(z))$ term).

\subsection{BiGANs}

Bidirectional GANs, first proposed by \cite{donahue2016adversarial}, include an encoder (E) that learns the inverse of the generator.  While the generator will learn a mapping from the latent dimension to data, the encoder will learn a mapping from data to the latent dimension.  The discriminator then must classify pairs of the form $(G(z), z)$ or $(x, E(x))$ as real or synthetic, where $z$ is noise from a standard distribution and $x$ is real data. 

\begin{align*}
    \max\limits_{D}\min\limits_{G, E} V(D, G, E)& =\\\mathbb{E}_{x\sim p_{data}(x)}&\big[\mathbb{E}_{z\sim p_E(\cdot|x)}[\log D(x,z)]\big]\\ + \mathbb{E}_{z\sim p_z(z)}&\big[\mathbb{E}_{x\sim p_G(\cdot|z)}[\log (1-D(x,z))]\big]
\end{align*}

\subsection{Anomaly Detection with BiGANs}

As first proposed in the work by \cite{schlegl2017unsupervised}, and further developed by \cite{zenati2018efficient}, variants of GANs that also learn an inverse of the generator can be used to detect anomalous data. Specifically, after training a generator, discriminator, and encoder, an anomaly score can be calculated for each data sample, where a higher score indicates greater likelihood of belonging to the anomalous class. In the current state-of-the-art GAN based anomaly detection \cite{zenati2018efficient}, a combination of a reconstruction loss $L_G$ and discriminator-based loss $L_D$ is used to determine the anomaly score, 
\begin{equation}
A(x)=\alpha L_G(x)+(1-\alpha)L_D(x), \label{score_egbad}
\end{equation}
where the reconstruction loss is given by $L_G(x)=\|x-G(E(x))\|$ and the discriminator loss is given by the cross-entropy, $L_D(x)=\sigma(D(x,E(x)),1)$.  We refer to this algorithm as EGBAD (Efficient GAN based anomaly detection) and in \cite{zenati2018efficient} the method is shown to outperform a variety of deep unsupervised models including anogan, variational auto-encoders, and deep auto-encoder GMM.

One drawback of GAN based anomaly detection such as EGBAD and GANomaly is that they are not detecting anomalies in multimodal datasets (as we will show experimentally in the next section).  Our approach is therefore not to view the latent variable $z$ as a single model Gaussian, but as a mixute of several Gaussians with means and covariances ($\mu_i$,$\Sigma_i)$.  Outliers can then be detected using a multimodal Mahalanobis distance for the anomaly score.  In particular, for a new data point in latent space, $z=E(x)$, the Mahalanobis distance to cluster $i$ is given by,
\begin{equation}
D^i_M(z) = \sqrt{(z - \mu_i)^{T} \Sigma_i^{-1} (z - \mu_i)}\label{mahl}
\end{equation}

The anomaly score is then given by the minimum distance, $D(x)=\min_i D^i_M$.  We note that the Mahalanobis based anomaly score produces not only improved anomaly detection results, as will be seen in the next section, but also up to 4x faster inference time over the cross-entropy loss in Equation \ref{score_egbad}. 

\subsection{Infinite Gaussian Mixture Model}

Because our goal is end-to-end learning for anomaly detection, we use an infinite Gaussian mixture model (IGMM) \cite{igmm} to automatically learn the number of clusters as well as the cluster means and covariances $(\mu_i,\Sigma_i)$ for the anomaly score defined by Equation \ref{mahl}.

IGMM \cite{igmm} is a Dirichlet Process Mixture Model in which the number of components can grow arbitrarily as data allows, hence the name Infinite Gaussian Mixture Model. IGMM assumes each cluster is modeled by a single Gaussian component and the base Dirichlet distribution serves as a prior for the parameters of these components (cluster mean $\mu$ and cluster covariance $\Sigma$).  As the name Gaussian mixture suggests, the bi-variate prior, $H$, involves a Gaussian prior over mean vectors and Inverse-Wishart over covariance matrices. More precisely H can be written as follows, 
\begin{eqnarray}
H = N(\mu | \mu_0, \Sigma_0 \kappa_{0}^{-1})W^{-1}(\Sigma|\Sigma_0, m)\label{eq:prior}
\end{eqnarray}
where $\mu_0$ is the mean of Gaussian prior, $\kappa_0$ is scaling constant that adjusts the dispersion of cluster center and parameter $m$ dictates the expected shapes of clusters. 
Note that Normal and  Inverse-Wishart distributions are conjugate, thus the posterior predictive distribution can be analytically derived, in the form of a multivariate Student-t distribution, by integrating out the component parameters $\{\mu_i, \Sigma_i\}$.  For inference we utilize Collapsed Gibbs Sampling \cite{igmm} due to the conjugacy between the model (Gaussian) and the prior (NIW). The generative model is illustrated in (\ref{eq:igmm})
\begin{eqnarray}
z_i & \sim & N(z_i | \mu_i, \Sigma_i) \nonumber\\
\{\mu_i, \Sigma_i \} & \sim & G \nonumber\\
G & \sim & DP(\alpha H) \label{eq:igmm}
\end{eqnarray}
where $H$ is defined by Equation (\ref{eq:prior}), $z_i$ is the data point from cluster $i$ and $\alpha$ is the concentration parameter of the Dirichlet process.

\section{Experimental results}

\subsection{Datasets}

In this section we run several experiments in order to compare our IGMM-GAN to several benchmark GAN based anomaly detection algorithms, as well as to evaluate the ability of GANs to generate realistic GPS trajectory data.  Our experiments are run on the following datasets.

\subsubsection{GeoLife GPS Trajectories}

The Geolife GPS trajectory dataset \cite{zheng2009mining,zheng2008understanding,zheng2010geolife}, consists of data collected from 180 users going about their daily routine movements.  The data consisted of time-series in the form of latitude, longitude, and velocity sampled every few seconds from each user.  A subset of the trajectory data had information on the mode of transportation and for our experiments we used only those trajectories that could be confirmed as ``car''.  As in other GAN based anomaly detection studies, we use a held-out class (specific human subject) during training to define an anomalous class in testing (and then evaluate over different choices of the subject held out).

\subsubsection{San Francisco Cabspotting}

Similar to Geolife, the SF Cabspotting data set consists of GPS trajectories from different cab drivers in San Francisco \cite{comsnets09piorkowski}.  This data consisted of latitudes, longitudes, and times, which were then expanded to include latitude, longitude, latitudinal velocity, and longitudinal velocity.  Cab data was split into routes by taking a set trip size of points.  Anomalous routes were found by training on a subset and finding the top anomaly scores.   

\subsubsection{MNIST}

Previous GAN based anomaly detection studies have used MNIST (a dataset of handwritten numbers) \cite{lecun2010mnist} for bench-marking competing methods.  Anomalies are defined by leaving out a digit from training and assessing the AUC (or other classification metric) of the anomaly score on a test data set which includes the held-out digit. 

\subsection{Architecture}

As mentioned previously, our model architecture is based on that of the BiGAN in \cite{donahue2016adversarial} and \cite{zenati2018efficient}. The arichetecture for the model is given in Table \ref{tab:my_label}.  The encoder consists of an input layer taking in a $4xN$ time series of latitude, longitude, latitude velocity and longitude velocity (or an $NxN$ image in the case of MNIST).  For the mobility data we used time series of trip segments of length $N=32$.  Wherease the encoder consists of several convolution and dense layers, the generator makes use of convolution transpose layers to facilitate learning of the inverse of the encoder.  The 2D convolution layers in the model are each followed by batch normalization and "Leaky ReLu" activation.  The discriminator is slightly more complex, beginning as two separate models, one composed similarly to the encoder which takes the real and generated data as input, and one containing dense layers which takes the latent representation as input. These two networks are then concatenated, ending in two final dense layers and a sigmoid activation.

\vspace{5mm}
\begin{table}[]
\centering
\begin{tabular}{l l l l l}
Layer & Units & BN & Activation & Kernel \\ 
\hline
$E(x)$ & & & & \\
Dense & 768 & & ReLU & \\
Convolution & 32 & \checkmark & ReLU & $3 \times 2$ \\
Convolution & 64 & \checkmark & ReLU & $3 \times 2$ \\
Convolution & 128 & \checkmark & ReLU & $3 \times 2$ \\
Dense & 100 & & & \\

$G(z)$ & & & & \\
Conv. Transpose & 128 & \checkmark & ReLU & $3 \times 2$ \\
Conv. Transpose & 64 & \checkmark & ReLU & $3 \times 2$ \\
Conv. Transpose & 32 & & & $3 \times 2$ \\
Dense & 1 & & Linear &  \\

$D(x)$ & & & & \\
Convolution & 64 & & Leaky ReLU & $3 \times 2$ \\
Convolution & 64 & \checkmark & Leaky ReLU & $3 \times 1$ \\
$D(z)$ & & & & \\
Dense & 512 & & Leaky ReLU &  \\
Concatenate & & & & \\
$D(x,z)$ & & & & \\
Dense & 1 & & Leaky ReLU &  \\
\end{tabular}
\caption{The architecture for our model, layer by layer. Units refer to number of filters in the case of convolution layers, and BN is Batch Normalization abbreviated.}
\label{tab:my_label}
\end{table}

Furthermore, combining the ideas from \cite{akcay2018ganomaly} and \cite{donahue2016adversarial}, we add onto the existing architecture a reconstruction loss term, taking into account the ability of the encoder and generator to reproduce a real image. This loss term helps ensure that not only can the generator's images fool the discriminator, but also that the encoder and generator function as closely as possible to inverses of one another. This loss is defined as:
\begin{align*}
 L_R = ||x - G(E(x))||_2
\end{align*}
We use an Adam optimizer \cite{kingma2014adam} with a learning rate of $lr = 1e^{-5}$ and $\beta = 0.5$. These parameters are sufficient for the generator and discriminator loss for our model to converge, similarly to the other models.

\begin{figure*}[h]
\centering
\includegraphics[width=.3\textwidth]{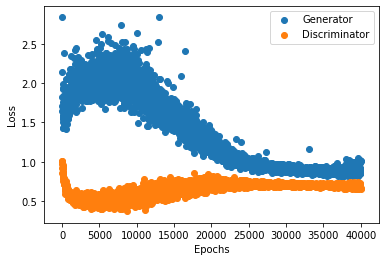}\includegraphics[width=.3\textwidth]{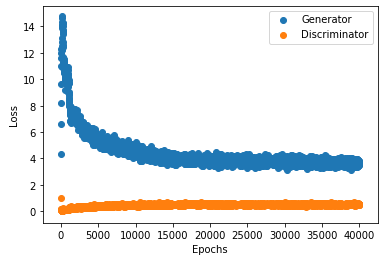}\includegraphics[width=.3\textwidth]{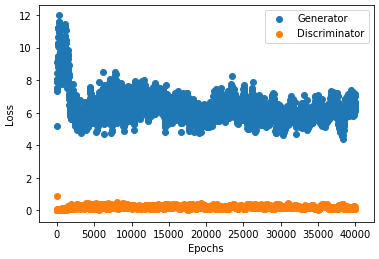}
\caption{Generator and discriminator loss by epoch for MNIST, digit 9 (Order: Ganomaly, EGBAD, Our Method)}
\label{fig:figure3}
\end{figure*}

\subsection{Hyper-parameter tuning for IGMM}

The hyper-parameters of IGMM are coarsely tuned to maximize the macro-f1 score. As the data is not well balanced, macro-f1 was chosen to suppress the dominance of large classes. IGMM has 4 hyper-parameters, $\{\kappa_0, m, \mu_0, \Sigma_0\}$ to be tuned. To simplify the tuning process, the prior mean, $\mu_0$, is set to the mean of data and we set $\Sigma_0$ to an identity matrix scaled by a parameter $s$. This left us with 3 parameters, $\{\kappa_0, m, s\}$ to tune. Parameter ranges and best triples are illustrated in Table \ref{tab:tuning}. The number of sweeps in the inference is fixed at 500, with 300 used for the burn-in period. Label samples are collected in every 50 iteration after burn-in and aligned by the Hungarian method to render final cluster labels. 

\begin{table}[htp]
\begin{subtable}{0.45\textwidth}
\centering
\begin{tabular}{|p{1cm}|p{5cm}|}
 \hline
 \textbf{HP}& \textbf{Range}  \\
 \hline \hline
 $\kappa_0$ & $0.01; 0.1; 1; 10; 100$\\  
 $m$ & $d+10; d+15; d+20; 5d; 10d; 100d$\\ 
 $s$ & $1; 3; 5; 7; 9$ \\ \hline 
\end{tabular}
\caption{Parameter ranges used in tuning}
\end{subtable}
\begin{subtable}{0.45\textwidth}
\centering
\begin{tabular}{|p{1cm}|p{2.3cm}|p{2.3cm}|}
 \hline
 \textbf{HP}& \textbf{MNIST} & \textbf{Geolife} \\
 \hline \hline
 $\kappa_0$ & $0.1$ & $0.1$\\  
 $m$ & $d+20$ & $d+15$\\ 
 $s$ & $7$ & $5$ \\ \hline 
\end{tabular}
\caption{Best triples from tuning}
\end{subtable}
\caption{Ranges for tuning and best triples used in experiments. HP stands for hyper-parameters}
\label{tab:tuning}
\end{table}

We restricted created clusters to ones with more than 50 points as IGMM may generate artificial small clusters to fit in distribution.

\subsection{Determining Anomaly Scores}

\begin{figure}
\includegraphics[scale=0.55]{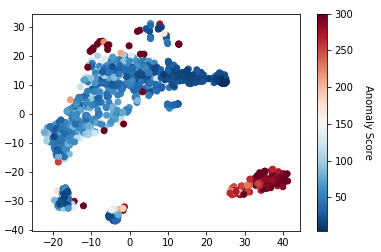}
\captionof{figure}{TSNE visualization of the latent dimension with anomaly scores from Geolife data}
\label{tsne}
\end{figure}

Anomaly scores were determined by using IGMM on the encoded training data to determine the cluster means and covariance matrices.  From there, an anomaly score was determined by the Mahalanobis distance to the nearest cluster. Figure \ref{tsne} shows an example TSNE visualization \cite{maaten2008visualizing} of the GeoLife data in latent space colored by the anomaly scores.  In this case, the driver that was not included in the training data is a clear outlying cluster reflected by the anomaly scores.  In the Geolife experimental section below we provide a detailed analysis of the performance of IGMM-GAN for detecting hidden drivers against several benchmark models.

\begin{figure}[h]
\includegraphics[scale=0.4]{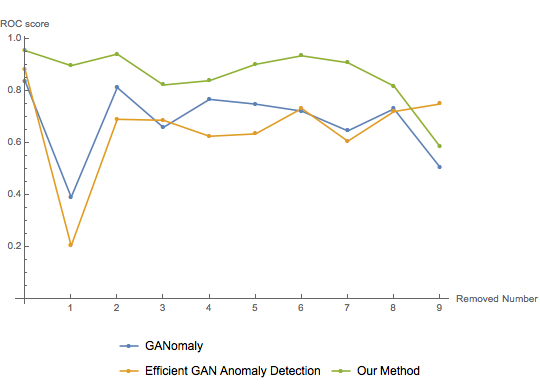}
\captionof{figure}{ROC AUC scores with MNIST data}
\label{mnist_auc}
\end{figure}

\subsection{Improving MNIST Benchmarks with IGMM}
Following the approach of \cite{zenati2018efficient}, we start by designating one digit as an anomalous class and remove it from the training dataset. For the remaining data we perform an 80/20 split into training and test sets and train the models for 40,000 epochs (where each epoch involves training on a random batch of 128 images). We repeat this process for each digit and for each anomaly detection method, scoring each method on its ROC AUC score. In Figure \ref{mnist_auc} we display the AUC scores of IGMM-GAN against GANomaly and EGBAD for each digit held out of testing.  The IGMM-GAN significantly improves the AUC scores for the majority of digits held out, especially for digits 1 and 7.

\subsection{San Francisco Cabspotting Experiment}
\begin{figure}[h!]
\centering
\includegraphics[scale=0.3]{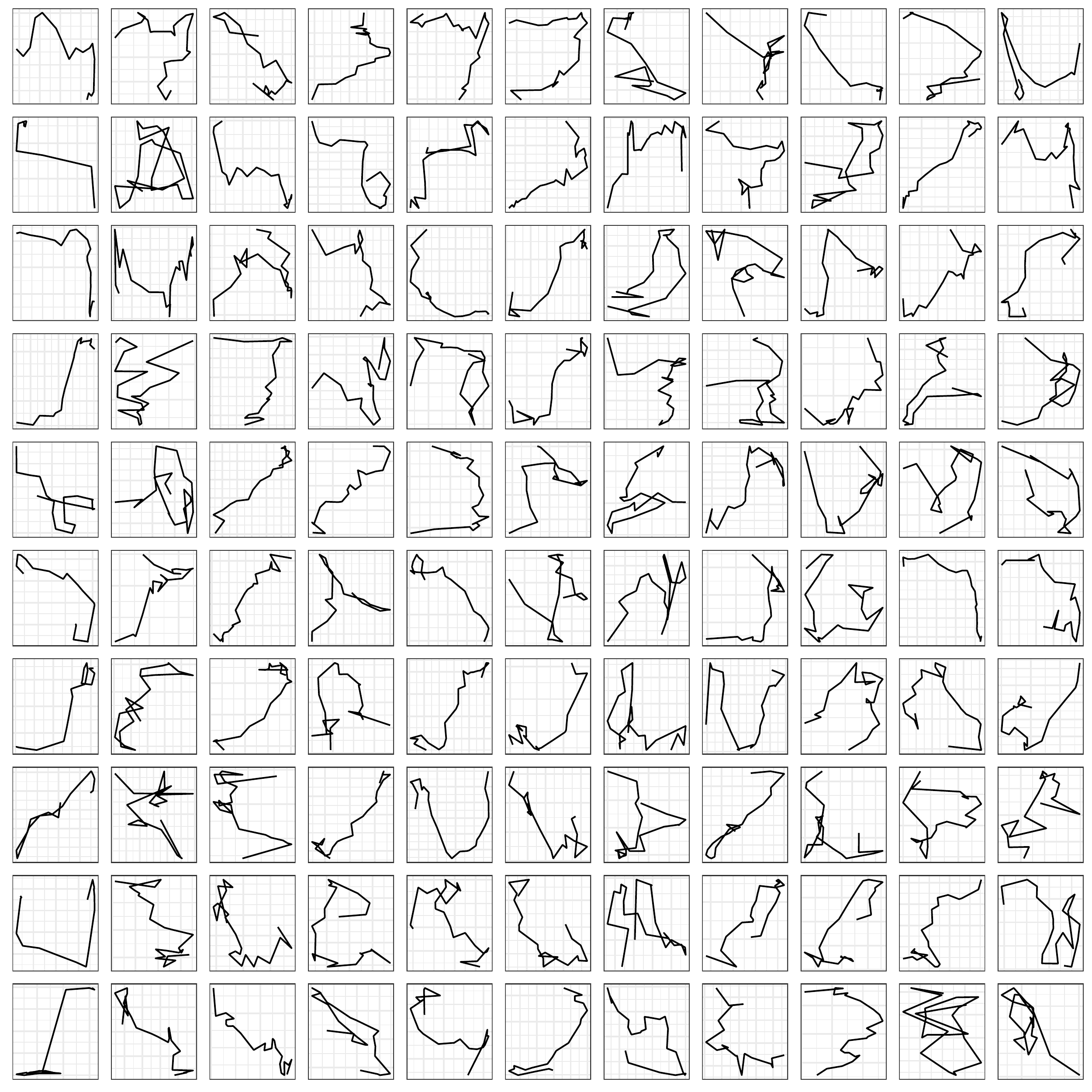}
\caption{Real and generated cab spotting routes.  Column 1: Real routes, Column 2-11: 10 closest generated routes (by Euclidean distance in the latent dimension).}
\label{cabroutes_lat}
\end{figure}

\begin{figure}
\centering
\includegraphics[scale=0.3]{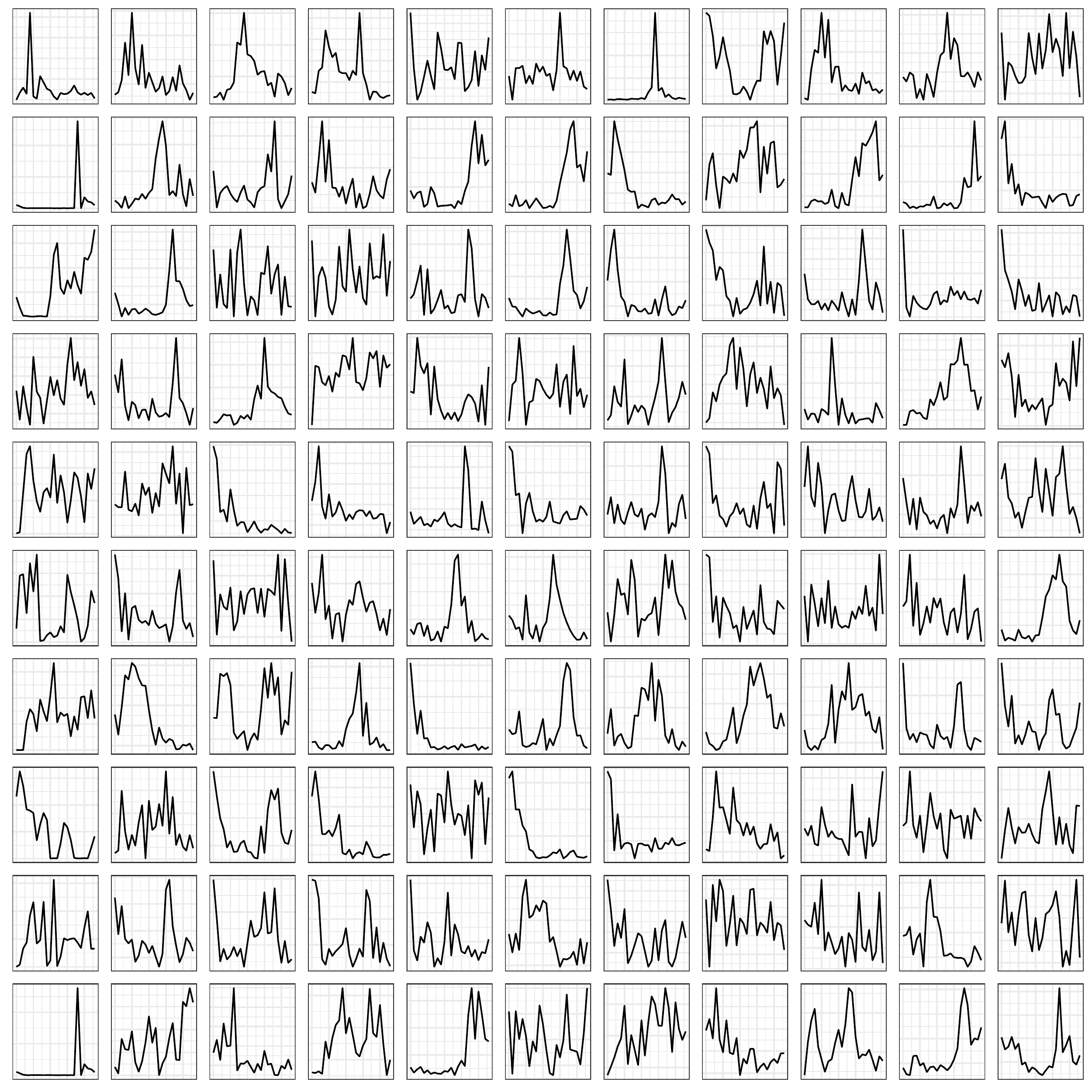}
\caption{Real and generated cab spotting speed profiles. Column 1: Real speed profiles, Column 2-11: 10 closest generated profiles (by Euclidean distance in the latent dimension).}
\label{cabroutes_vel}
\end{figure}

\subsubsection{Generating Routes}
We first examine the IGMM-GAN's ability to generate realistic synthetic trajectories.  Note that the routes were created using a generator trained on both latitude and longitude as well as latitudinal and longitudinal velocity.  Including the velocities gave slightly better results, as the sampling interval in the data was not uniform, hence with the added velocities, the generator could infer the time step.

We display 10 example real trajectories (Figure \ref{cabroutes_lat}) and speed profiles (Figure \ref{cabroutes_vel}) in the left most column of each grid along with the 10 closest trajectories and speed profiles (closest in latent space) to each real route.  The generated trajectory data in Figure \ref{cabroutes_lat} matches the real data both upon visual inspection and in terms of convergence of the discriminator and generator loss.  Qualitatively, both the synthetic and real routes, for latitude and longitude, are characterized by long stretches of straight lines or slight curves followed by segments of turns and short paths, as seen in Figure \ref{cabroutes_lat}.  The velocities in the generated routes also follow a natural pattern of increase and decrease, as seen in Figure \ref{cabroutes_vel}. The generated velocities also follow general traffic patterns with longer distances between points often having higher velocity and slower velocities often being placed near the center of the city.


\subsubsection{Finding Anomalous Routes}

Using the above-mentioned methods, we assigned an anomaly score to each route in the Cabspotting dataset, leading us to find several types of anomalies.  The first category of anomalous routes were those with GPS noise: in the dataset certain trips contain inaccurate points where the coordinates may have noise on the order of miles.  Using both latitude, longitude and the respective velocities, these routes gave the highest anomaly scores.  We therefore removed these trips prior to training.  With the GPS noise trajectories removed, different types of anomalies emerged.  The anomalies found were generally in two categories, anomalies in the route taken and anomalies in velocity.


\begin{figure}
\centering
\includegraphics[scale=0.5]{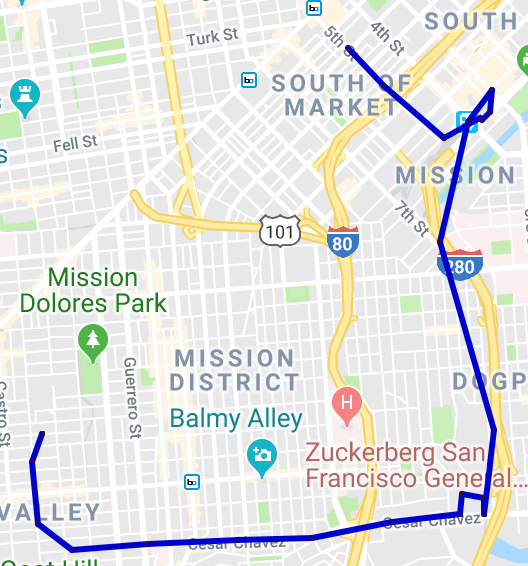}
\captionof{figure}{Anomalous route from cabspotting data}
\label{cabanom1}
\end{figure}

Figure \ref{cabanom1} is an example of the route anomaly.  Note that in the bottom right corner the taxi takes a detour from the usual route.  Other anomalies in the route taken included routes that had a large number of turns or ones that were right next to major roads, but did not use them.

\begin{figure}[h]
\centering
\includegraphics[scale=0.4]{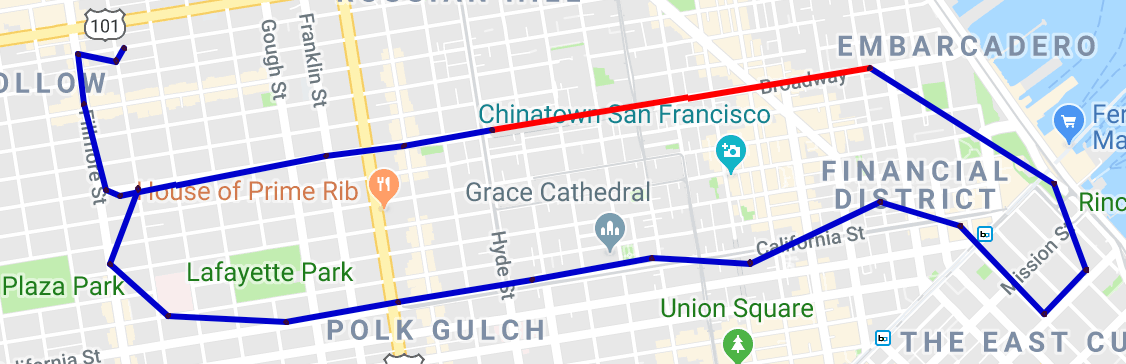}
\captionof{figure}{Anomalous route from cabspotting data, note the red section is much faster than the blue}
\label{cabanom2}
\end{figure}

In Figure \ref{cabanom2}, the red part shows where the speed, instead of a normal speed for the area, was much higher (closer to highway speeds).  Anomalies in the velocity generally were either speeds that were too fast for the area, such as in the image above, or speeds that were too slow for the area, such as driving below 30mph on a highway.  Sometimes velocity anomalies were those where the acceleration suddenly changed and one segment would be slow and the next much faster.  Thus the IGMM-GAN is able to detect anomalous routes, but also driver behaviors with the same model (using an end-to-end learning  framework).

\subsection{Geolife Hidden Driver Experiment}

\begin{figure}[h]
\includegraphics[scale=0.4]{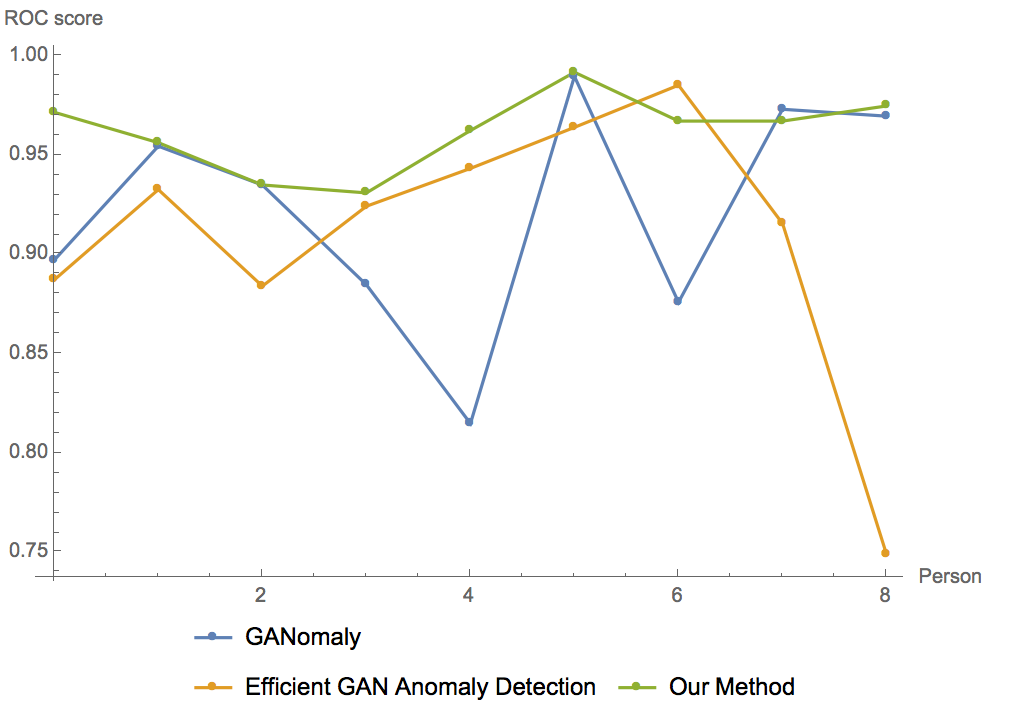}
\captionof{figure}{ROC AUC Scores for Different Drivers from GeoLife Data}
\label{auc_geo}
\end{figure}

Our next experiment, with the GeoLife data, focuses on distinguishing the emergence of a new driver not contained in the training dataset.  Here a similar method to the MNIST experiment was followed: one driver was left out of training data, then included in the testing data as a separate class.  In Figure \ref{auc_geo}, we compare the three methods from before on each of nine held out drivers.  Here we see that GANomaly and EGBAD perform better on some drivers and worse on others compared to each other.  However, IGMM-GAN is consistently at least as accurate as both methods and in 6 cases it shows the best AUC scores.  

\section{Conclusion}

We have made three main contributions in this paper: 
\begin{itemize}
    \item 
    We improved GAN based anomaly detection through the introduction of the multimodal IGMM-GAN.
    \item
    We showed that GAN can be successfully applied to generating synthetic human trajectories.  Synthetic data sets of this type will be useful as ground truth to benchmark existing anomaly detection methods for human mobility and can also aid in data augmentation.
    \item
    We showed that the IGMM-GAN can find route-based anomalies, anomalous driver behavior, and detect hidden drivers.
\end{itemize}
We believe that the IGMM-GAN will serve as a complimentary tool to existing algorithms for anomaly detection in human mobility data that require spatial grids and feature engineering. Our method may also find application to anomaly detection in other domains where the data is multimodal.

\section{Acknowledgements}

This work was supported in part by NSF grants ATD-1737996, REU-1343123, and SCC-1737585. 

\bibliographystyle{aaai}
\bibliography{sources}

\end{document}